\documentclass{article}

\usepackage{PRIMEarxiv}

\usepackage[utf8]{inputenc} 
\usepackage[T1]{fontenc}    
\usepackage{hyperref}       
\usepackage{url}            
\usepackage{booktabs}       
\usepackage{amsfonts}       
\usepackage{nicefrac}       
\usepackage{microtype}      
\usepackage{lipsum}
\usepackage{fancyhdr}       
\usepackage{graphicx}       
\graphicspath{{media/}}     
\usepackage{hyperref}
\usepackage{url}
\usepackage[utf8]{inputenc} 
\usepackage[T1]{fontenc}    
\usepackage{hyperref}       
\usepackage{url}            
\usepackage{booktabs}       
\usepackage{amsfonts}       
\usepackage{nicefrac}       
\usepackage{microtype}      
\usepackage{xcolor}         
\usepackage{graphicx}
\usepackage{subcaption}
\usepackage{amssymb}
\usepackage{tikz}
\usepackage{color}
\usepackage{amsmath}
\usepackage{arydshln}
\usepackage{wasysym}
\usepackage{colortbl}
\title{Attention Head Purification: A New Perspective to Harness CLIP for Domain Generalization
}

\author{
  Yingfan Wang \\
  Beihang University \\
   \And
  Guoliang Kang \\
  Beihang University \\
}

\begin{document}
\maketitle

\begin{abstract}
Domain Generalization (DG) aims to learn a model from multiple source domains to achieve satisfactory performance on unseen target domains. Recent works introduce CLIP to DG tasks due to its superior image-text alignment and zeros-shot performance. 
Previous methods either utilize full fine-tuning or prompt-learning paradigms to harness CLIP for DG tasks. Those works focus on avoiding catastrophic forgetting of the original knowledge encoded in CLIP but ignore that the knowledge encoded in CLIP in nature may contain domain-specific cues that constrain its domain generalization performance. 
In this paper, we propose a new perspective to harness CLIP for DG, \emph{i.e.,}
attention head purification. We observe that different attention heads may encode different properties of an image and selecting heads appropriately may yield remarkable performance improvement across domains. Based on such observations, we purify the attention heads of CLIP from two levels, including \textit{task-level purification} and \textit{domain-level purification}.
For task-level purification, we design head-aware LoRA to make each head more adapted to the task we considered. 
For domain-level purification, we perform head selection via a simple gating strategy. 
We utilize MMD loss to encourage masked head features to be more domain-invariant to emphasize more generalizable properties/heads.
During training, we jointly perform task-level purification and domain-level purification.
We conduct experiments on various representative DG benchmarks. Though simple, extensive experiments demonstrate that our method performs favorably against previous state-of-the-arts.
\end{abstract}


\section{Introduction}
Deep learning has attained remarkable success on various downstream tasks in computer vision, typically under the assumption that both training and test samples are identically distributed. 
However, in practice, test data distributions (target) are usually different from the training ones (source). 
In such cases, the performance of deep neural networks may degenerate severely on target, showing a poor domain generalization ability. 
To mitigate the domain shift, a series of Domain Generalization (DG) methods~\cite{1,5,12,13,15,18,mix-style} are proposed to transfer the knowledge learned from multiple source domains to unseen target domains, via domain-invariant learning~\cite{3,4,kl1,adv2,adv3,kl2}, meta-learning techniques~\cite{meta1,metareg,meta2,15}, or specifically-designed data augmentations~\cite{advaug,aug1,aug2}. 
The remarkable progress achieved by those works can be largely attributed to a well-initialized feature extractor provided by ImageNet-pretrained~\cite{imgnet} backbones.
Recent proposed Contrastive Language–Image Pre-training (\emph{i.e.,} CLIP~\cite{itpairs1}) learns from large amounts of image-text pairs~\cite{itpairs2} and demonstrates impressive zero-shot learning performance. 
It may potentially serve as a better foundation for mitigating the performance gap across domains for specific downstream tasks. 

Despite superior zero-shot classification performance of CLIP, it is not trivial to harness CLIP for specific domain generalization tasks beyond its zero-shot classification ability. 
Directly fine-tuning CLIP may obtain even worse performance than zero-shot classification \cite{cost1,cost2}.
Previous works tackle this issue mainly by introducing various regularizations to avoid the forgetting of original knowledge encoded in CLIP. 
For example, ~\cite{clipood} fully fine-tunes CLIP's image encoder with modified contrastive loss to mitigate overfitting and proposes beta moving average to perform a temporal ensembling along the fine-tuning trajectory. ~\cite{5} proposes mutual information regularization between the original CLIP and the fine-tuned one to prevent the two models from deviating too much. 
Although those works achieve improved domain generalization performance, a natural question arises: \textit{is the way to avoid knowledge forgetting sufficient to harness CLIP for domain generalization?}

\begin{figure}
    \centering
    \includegraphics[width=1\textwidth]{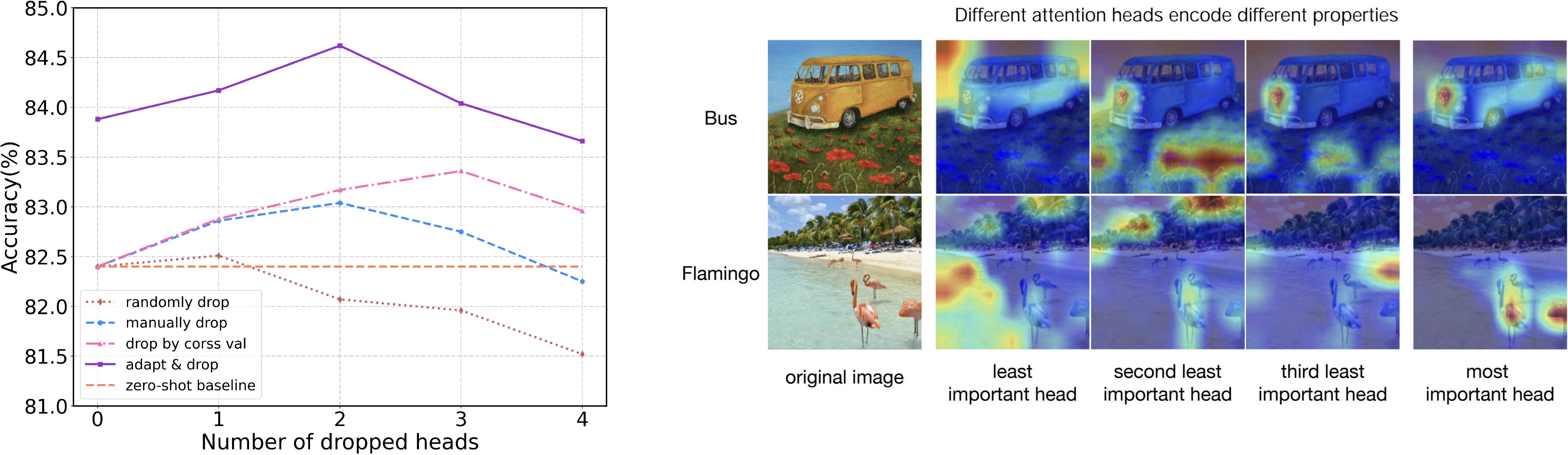}
    \caption{\textbf{Left}: We use different strategies to evaluate the importance of each attention head on domain generalization and drop the least important ones to see how accuracy changes. 
    The strategies we adopt include ``randomly drop(\textcolor{brown}{--$\lozenge$--})'', ``manually drop(\textcolor{cyan}{--\(\CIRCLE\)--})'', ``drop by cross-validation(\textcolor{magenta}{--\(\blacktriangle\)--})'', and ``adapt \& drop(\textcolor{violet}{--\(\blacksquare\)--})''. 
    Details of strategies can be found in Appendix.
    Via appropriately dropping heads, CLIP's domain generalization performance can be improved. 
    \textbf{Right:} Attention map~\cite{attentionmap} generated by specific heads. The middle columns are from least important heads determined by the cross-validation strategy. They all capture lots of background information. The last column represents the most important one whose attention map mainly focuses on the object itself. 
    The experiments are conducted on OfficeHome~\cite{officehome}. Best viewed in color.}
    \label{fig-overview}
    \vspace{-5mm}
\end{figure}

Recent work~\cite{interpret} shows that different attention heads of CLIP's image encoder (ViT-based) may encode different properties of an image. Inspired by this work, we conduct ablation experiments to verify the effectiveness of each head in the context of domain generalization. Specifically, we utilize different ways to evaluate the importance of attention heads on domain generalization (\emph{e.g.,} manual evaluation, 
evaluation by cross-validation, 
\emph{etc.}) 
and drop the least important ones.
As shown in Figure~\ref{fig-overview}(a), we observe that via appropriately dropping some attention heads, we may achieve a much better domain generalization performance. \textcolor{black}{This phenomenon implies that not all attention heads are domain generalizable and some heads may harm the domain generalization ability of CLIP, which means those heads may contain non-generalizable cues (\emph{i.e.,} domain-specific cues)}. To further illustrate our findings, we provide attention visualizations in Figure~\ref{fig-overview}(b). We observe that the attention maps~\cite{attentionmap} of the least important heads mainly highlight task-irrelevant regions like background, while the most important head pays more attention to the object itself.
As a result, simply avoiding the knowledge forgetting of CLIP may not be optimal for domain generalization tasks. We need to purify the attention heads to make CLIP more task-adapted and domain-generalizable.

In this paper, we propose a new perspective to harness CLIP for domain generalization tasks, \emph{i.e.,} through attention head purification. Specifically, we perform two kinds of attention head purification during training, which are named \textit{task-level purification} and \textit{domain-level purification}. 
For task-level purification, we aim to purify the attention head of CLIP to maintain more task-related knowledge. Technically, we adopt LoRA to realize this goal. Different from conventional LoRA implementation, we design head-aware LoRA (HA-LoRA) to purify and adapt each head more accordingly.
For domain-level purification, we aim to purify the attention heads of CLIP to make the resulting features more invariant across domains. 
We design a simple learnable gating strategy to select the heads that most benefit the domain generalization performance. To realize this, in addition to the cross-entropy loss on source images, we utilize Maximize Mean Discrepancy (MMD) loss to encourage the gates to emphasize more domain-invariant head features. 
Note that we do not utilize the gradients of MMD loss to update the HA-LoRA. 
In this way, we may decouple task-level and domain-level purification to an extent, 
\emph{i.e.,} we expect HA-LoRA to focus only on encoding rich \textit{task-related} properties, while leaving the goal of selecting 
domain-invariant properties to domain-level purification.
During training, we jointly train the head-aware LoRA and the gates of attention heads to perform the task and domain-level purification simultaneously. Pre-trained parameters in the image encoder and the text encoder of CLIP are frozen throughout the training. 
We conduct extensive experiments on various representative domain generalization benchmarks including Office-Home~\cite{officehome}, DomainNet~\cite{domainnet}, PACS~\cite{pacs}, VLCS~\cite{vlcs} and TerraIncognita~\cite{ti}.
All those experiments verify the superiority of our proposed method. 

In a nutshell, our contributions can be summarized as \textbf{1)} We observe that not all attention heads of CLIP are domain generalizable in terms of specific tasks and propose a new perspective to harness CLIP for DG through attention head purification, \emph{i.e.,} optimizing attention heads to make them more task-adapted and domain generalizable. 
\textbf{2)} We decouple the attention head purification from two levels including task-level purification and domain-level purification. We design head-aware LoRA and learnable gating strategy to perform the two kinds of purification respectively. 
\textbf{3)} Extensive experiments on representative domain generalization benchmarks demonstrate that our method performs favourably against the previous state-of-the-art.  

\section{Related works}
\textbf{Vision-Language Pre-training.} Vision-Language models(VLMs) connect images and texts through a common embedding space to enable cross-modal learning~\cite{devise,socher2013zero, write}. Recent advances employ architectures with better representation learning abilities such as Transformer~\cite{attention} and webscale training datasets and build stronger vision-language pre-trained models. One type of approach learns the common embedding space by masked language modeling or masked region prediction~\cite{lxmert,vl,kim2021vilt}. Another typical type of vision-language pretraining is contrastive image-language pre-training, such as CLIP~\cite{itpairs1} and ALIGN~\cite{itpairs2}. Recent research also seeks to improve the pre-training paradigm, such as using additional supervision~\cite{li2021supervision,slip}, employing pre-trained image encoders~\cite{zhai2022lit}, and adding cross-modal and in-modal consistency constraints~\cite{cyclip}. In this paper, instead of designing better pretraining techniques, we aim at utilizing recent advances in vision-language pre-trained models such as CLIP and achieving better generalization performance.

\textbf{Domain Generalization(DG).} Domain generalization aims to learn generalized representations from multiple source domains that can generalize well on arbitrary unseen target domains. Most DG methods perform domain alignment ~\cite{dann,4,adv2,adv3,kl1,kl2,mmd}, to learn domain invariant features by reducing the distance of distributions across multiple domains. Specifically, to achieve domain-invariance, ~\cite{4,adv2,adv3,mmd,dann} imply adversarial learning, ~\cite{mmd} minimizes maximum mean discrepancy (MMD), and ~\cite{meta1,meta2,meta3} utilize meta-learning techniques. In addition, ~\cite{img-level1,img-level2,dsu,styleneophile,mix-style,padain,ccfp,aloft} use domain augmentation to enrich the style diversity of source data at image level or feature level.
The remarkable progress achieved by those works can be largely attributed to a well-initialized feature extractor provided by ImageNet pre-training~\cite{imgnet}.
Recent proposed Vision-language pre-trained models such as CLIP exhibit impressive zero-shot generalization ability. Its zero-shot performance exceeds the aforementioned methods in various DG tasks. 
In this paper, we focus on leveraging CLIP to further improve domain generalization performance.

\textbf{CLIP-based Domain Generalization.} Despite the good zero-shot performance, research finds that directly finetuning CLIP models with task-specific data will damage the alignment in joint vision-language spaces~\cite{diagnosing} and harm CLIP's domain generalization ability~\cite{itpairs1,cost2}. Recent advances explore adapting CLIP to downstream tasks without affecting its generalization ability by adapter learning~\cite{clip-adapter,tip}, model ensemble~\cite{cost2}, regularized fine-tuning~\cite{cost2,5,clipood,gestur}, distillation~\cite{vl2v,rise}, and prompt learning~\cite{promptstyler,coop,dpl,duprg,stylip,dspl,spg}.
Different from most existing works that use regularization to avoid knowledge forgetting of CLIP, we propose attention head purification to harness CLIP for DG.

\section{Method} \label{sec: method}
\vspace{-2mm}
\subsection{Preliminary}
\vspace{-2mm}
\textbf{Background.} We aim to tackle the domain generalization problem based on CLIP. 
Specifically, suppose we have multiple source domains for training. 
Our goal is to adapt CLIP with labeled samples from source domains to make it perform well on unseen target domains.
Usually, the distribution of target domain data is distinct from that of each source domain.

\textbf{CLIP Revisiting.} 
CLIP is a large-scale visual-language model that consists of an image encoder and a text encoder. 
CLIP is trained with 400 million web-scraped image-text pairs~\cite{itpairs1}.
The contrastive loss is imposed to encourage the features from paired image-text to be more aligned than unpaired ones.
As a result, CLIP is readily adopted to downstream tasks even without any fine-tuning, showing impressive zero-shot classification ability. Specifically, given $C$ classes with their class names, we may construct a text description for each class, \emph{e.g.,} ``A photo of a \#classname". For a specific image, we extract its image feature, and compare the extracted image feature with text features of different classes. Then, the class of text feature which has a highest cosine similarity with the image feature is viewed as the predicted label of the image.
Formally, the zero-shot classification process can be represented as
$\hat{y} = \mathrm{argmax}_c \mathrm{cos}(I,T_c)$, 
where $I$ denote the image feature, $T_c$ denotes the text feature of class $c\in\{0,1,2,\cdots, C-1\}$, and $\hat{y}$ is the predicted label for the image.

\noindent
\textbf{LoRA Revisiting.} Low-rank adaptation (LoRA)~\cite{lora} is one of the most popular parameter-efficient fine-tuning (PEFT) methods. It assumes that the changes of parameters lie in a low-rank space when the model is fine-tuned on a downstream task. 
Specifically, for a linear layer with the input dimension \(d_I\) and the output dimension
\(d_O\), we represent its weight with \(W^{d_O\times d_I}\). Then LoRA reparametrizes the pre-trained weight \(W\) by expanding a branch with two matrices, \( A \in\) \(R^{d_O \times r}\) and \(B \in R^{r \times d_I}\).
Typically, \(r\) is much smaller than the input dimension \(d_I\) and output dimension \(d_O\), making \(A\) a dimension increasing matrix and $B$ a dimension reduction matrix.
Finally, LoRA modifies the forward propagation in this linear layer as \(o = We + ABe\) where $e$ and $o$ denote the input and output features of this layer respectively. 
During adaptation to the downstream tasks, we freeze the
pre-trained weight \(W\) and only update \(A\) and \(B\).
\vspace{-2mm}
\subsection{Attention Head Purification of CLIP}
\vspace{-2mm}
\begin{figure}
    \centering
    \includegraphics[width=1\textwidth]{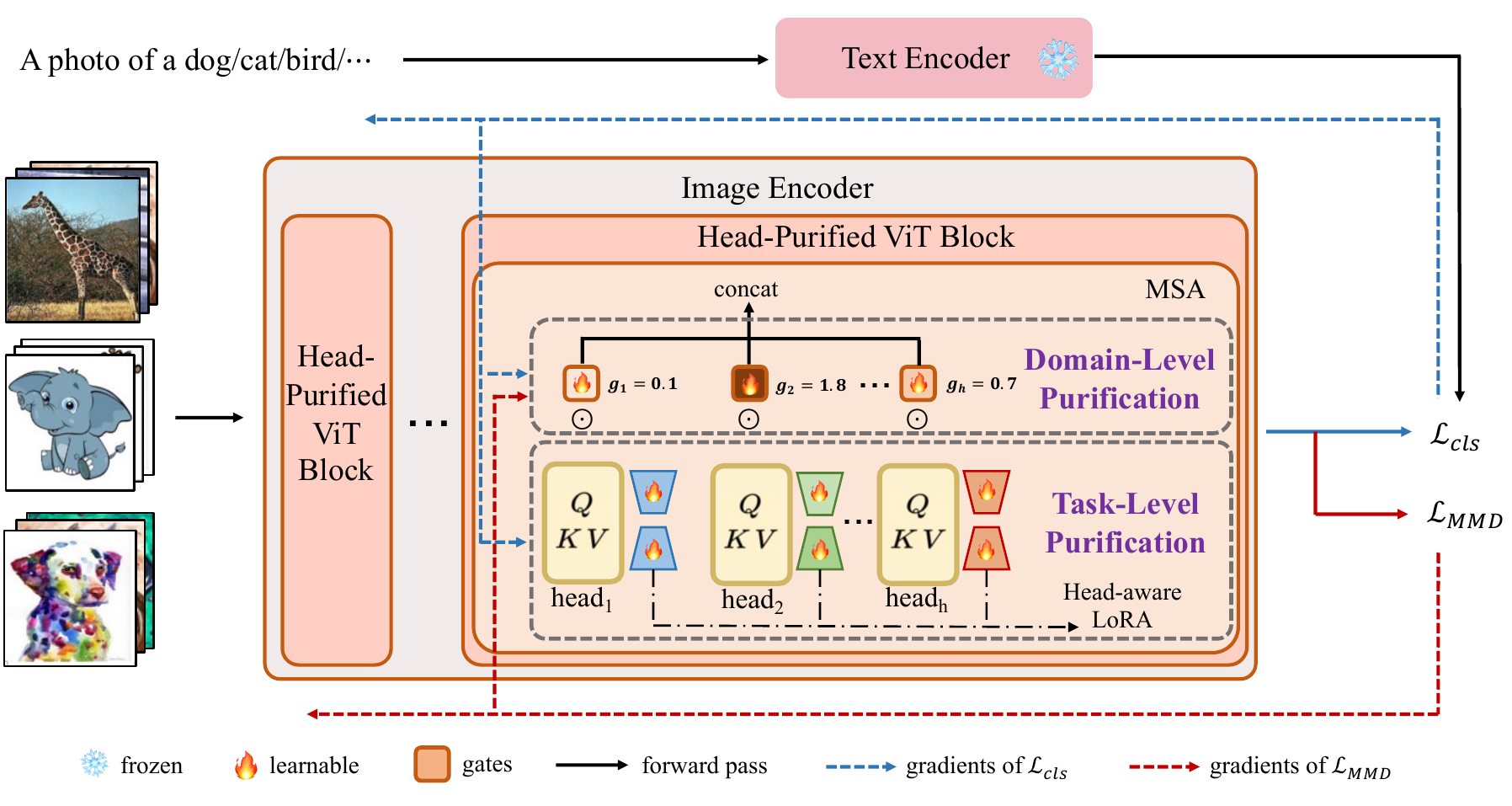}
    \caption{The architecture of Attention Head Purification. We design head-aware LoRA to perform task-level purification and domain-invariant gating strategy to perform domain-level purification. Further, we minimize \(\mathcal{L}_{cls}\) (Section~\ref{ha-lora}) to update head-aware LoRA and minimize \(\mathcal{L}_{cls},\mathcal{L}_{\mathrm{MMD}}\) (Section~\ref{domainpuri}) to update the gates of heads.} 
    \label{fig-pip}
    \vspace{-3mm}
\end{figure}

\textbf{Overall Framework.} 
The image encoder of CLIP encodes rich properties of an image into image features. For a specific downstream task, not all properties are beneficial, \emph{i.e.,} some properties may not be task-related, and some properties may not be domain-invariant. As shown in~\cite{interpret},  different attention heads may encode different properties of an image.  
From this point of view, we propose to purify the attention heads of CLIP from two levels, including task-level purification (detailed in Section~\ref{taskpuri}) and domain-level purification(detailed in Section~\ref{domainpuri}). 
For task-level purification, we purify the attention head of CLIP using our designed head-aware LoRA (HA-LoRA) to focus on task-related properties. 
For domain-level purification, we design a simple learnable gating strategy to emphasize generalizable heads while restraining domain-sensitive heads.
During training, we add HA-LoRA and head gating into certain layers of CLIP's image encoder as shown in Figure~\ref{fig-pip} to conduct task-level purification and domain-level purification simultaneously.


\vspace{-1mm}
\subsubsection{Task-Level Purification with Head-Aware LoRA} \label{taskpuri}
\vspace{-2mm}
From the task-level, we use LoRA to purify each attention head, \emph{i.e.,} encouraging the heads to focus on task-related patterns while ignoring task-irrelevant ones. Technically, for a linear projection \(W\), we may add a branch which sequentially multiplies a dimension reduction matrix \(B \in \mathbb{R}^{r \times d_I}\) and a dimension increasing matrix \(A \in \mathbb{R}^{d_O \times r}\). Then, the forward propagation of this linear layer is modified as
\begin{equation}
     o = We + ABe = W'e,
     \label{l_lora}
\end{equation}%
where \( W'= W + AB\) denotes the adapted weight, \(e\) denotes the input of the projection, and \(o\) denotes the output. 
We only use LoRA to adapt the linear projections of query ($Q$) and value ($V$) in multi-head self-attention (MSA) blocks.
To improve the effect of task-level purification and facilitate the following domain-level purification, we propose using head-aware LoRA instead of the conventional LoRA in our framework. 

\textbf{Head-aware LoRA.}\label{ha-lora}
For the pre-trained weight \( W \in \mathbb{R}^{d_O \times d_I} \) of the \(Q\) or \(V\) projection in a MSA block, we can split $d_O$ into $H$ groups (\emph{i.e.,} $W_1, W_2, \cdots, W_H$) where $H$ denotes the number of attention heads in a MSA block. As a result, $W_h\in\mathbb{R}^{n\times d_I}, h\in\{1,2,\cdots, H\}$ where $n=d_O/H$. 
We split the matrix $A$ in the same way and obtain $A_1, A_2, \cdots, A_H$ where $A_h \in \mathbb{R}^{n\times r}$.
Then the adapted weight $W'$ of the conventional LoRA can be reformulated as
\begin{equation}
        W' = W + AB 
           = W + \begin{pmatrix} A_1B \\ A_2B \\ \vdots \\A_HB \end{pmatrix} 
           = \begin{pmatrix} W_1+A_1B \\ W_2+A_2B \\ \vdots \\W_H+A_HB \end{pmatrix}. 
        \label{lora-group}
\end{equation}%
From Eq.~(\ref{lora-group}), we observe that in the conventional LoRA, different from \(A_h\) which is distinct with respect to different heads, $B$ is shared by all the heads. As a result, purifying one head may interfere with the other head, rendering the head purification less effective.   

To mitigate such interference between different heads, we propose head-aware LoRA, denoted as HA-LoRA. As shown in Task-Level Purification module in Figure~\ref{fig-pip}, we set independent \(B_h \in \mathbb{R}^{r \times d_I}\) for each head. The adapted weight \(W^{'}\) for HA-LoRA can be represented as:
\begin{equation}
         W' = W + \begin{pmatrix} A_1B_1 \\ A_2B_2 \\ \vdots \\A_HB_H \end{pmatrix}
        = \begin{pmatrix} W_1+A_1B_1 \\ W_2+A_2B_2 \\ \vdots \\W_H+A_HB_H \end{pmatrix} 
\end{equation}%
\textcolor{black}{
Different from~\cite{hydra}, which also uses independent \(B\) to handle different tasks, we use independent \(B\) to modulate different heads. The two approaches are technically similar but differ in their underlying motivations.}

We follow the vision-language contrastive learning strategy to update the parameters of HA-LoRA. For each class \(c\), we manually generate a text describing it, \emph{e.g.,} ``A photo of a \#classname''. We then get the text embedding of each class $T_c$ extracted by the text encoder. Then, we impose a contrastive loss between image features and the text features of different classes to encourage the image feature to be more aligned with the text feature of ground-truth label, \emph{i.e.,}
\begin{equation}
     \mathcal{L}_{cls} = -\frac{1}{N}\sum_{i=1}^{N}\mathrm{log} \frac{\mathrm{exp}(\mathrm{cos}(s_i,T_y)/\tau)}{\sum_{c=1}^C \mathrm{exp}(\mathrm{cos}(s_i,T_c)/\tau)}
     \label{loss_cls}
\end{equation}%
where $s_i$ denotes the image feature for the $i$-th sample, $T_c$ denotes the text feature of class $c$, $y$ is the ground-truth label for the $i$-th image and \(\tau\) is the temperature hyper-parameter. 
\vspace{-2mm}
\subsubsection{Domain-Level Purification with Domain-Invariant Gating}\label{domainpuri}
\vspace{-2mm}
By task-level purification, we can wipe off task-unrelated properties, making each head more adapted to current task. After this, we also need to perform domain-level purification, aiming to maintain or emphasize the most generalizable or invariant attention heads across domains. 
To realize this, we design a domain-invariant gating (DIG) scheme. 
Specifically, in a MSA block, we set a series of learnable gates $g_1, g_2, \cdots, g_H$, the number of which equals the number of attention heads $H$. To evaluate the relative importance of each head, we apply softmax operation to the gates, \emph{i.e.,}
$\hat{g}_1, \hat{g}_2, \cdots, \hat{g}_H = \text{Softmax}(g_1, g_2, \cdots, g_H)$.
Then, we apply the gates to the features of different heads which are the outputs of the scaled dot-product attention operation. We concatenate the gated features from all the heads and obtain $f^g$ as
\begin{align}\label{eq-gate}
f^g = \gamma [\hat{g}_1f_1, \hat{g}_2f_2, \cdots, \hat{g}_Hf_H]
\end{align}%
where $f_1, f_2, \cdots, f_H$ denote the features of different attention heads. The $\gamma$ equals to the number of heads, which is used to compensate the scale changes after softmax operation.

During training, in addition to \(\mathcal{L}_{cls}\), we also utilize Maximum Mean Discrepancy (MMD) loss~\cite{dan,jan} to measure the distribution discrepancy of features between different source domains, \emph{i.e.,}
\begin{equation}
           \mathrm{MMD}(S^p,S^q) = \frac{1}{N^2} \sum_{i=1}^{N}\sum_{i'=1}^{N}k(s^{p}_{i},s^{p}_{i'}) + \frac{1}{M^2} \sum_{j=1}^{M}\sum_{j'=1}^{M}k(s^{q}_{j},s^{q}_{j'}) \nonumber\\
           - \frac{2}{MN} \sum_{i=1}^{N}\sum_{j=1}^{M}k(s^{p}_{i},s^{q}_{j})
           \vspace{-2mm}
\end{equation}
\begin{equation}
           \mathcal{L}_{\mathrm{MMD}} = \frac{2}{d(d-1)} \sum_{p=1}^{d-1} \sum_{q=p+1}^{d} \mathrm{MMD}(S^p,S^q) 
\end{equation}
where \(S^{p} = \{s^{p}_{i}\}_{i=1}^N\) are image features of $p$-th source domain output by CLIP's image encoder, $N$ denotes the number of samples for domain $p$ within a mini-batch, \(d\) denotes the number of source domains and \(k\) denotes the Gaussian kernel~\cite{dan,jan}.

In this way, if the gates select/emphasize heads which are more generalizable, the resulting features are more generalizable, rendering $\mathcal{L}_{\mathrm{MMD}}$ small, otherwise the $\mathcal{L}_{\mathrm{MMD}}$ will be large. Thus, minimizing $\mathcal{L}_{\mathrm{MMD}}$ will encourage the gates to update towards selecting/emphasizing the most generalizable or domain-invariant heads.

\vspace{-2mm}
\subsection{Objective}\label{object}
\vspace{-2mm}
Since both task-level purification and domain-level purification contribute to CLIP's domain generalization performance, we want to combine them to achieve further performance improvement. We perform end-to-end training that simultaneously conducts task-level purification and domain-level purification as shown in Figure~\ref{fig-pip}.
The final optimization objective is as follows
\begin{equation}
     \underset {\theta_1, \theta_2}{\mathrm{min}} \mathcal{L}_{cls} + \underset {\theta_2}{\mathrm{min}} \ \alpha \mathcal{L}_{\mathrm{MMD}}, 
     \label{full_obj}
\end{equation}%
where $\theta_1=\{A_1, A_2, \cdots, A_H, B_1, B_2, \cdots, B_H\}$, and $\theta_2=\{g_1, g_2, \cdots, g_H\}$. The $\alpha$ controls the strength of MMD loss.

As shown in Eq.~(\ref{full_obj}), we do not use MMD loss to update HA-LoRA. 
In this way, we may decouple task-level and domain-level purification to an extent, 
\emph{i.e.,} we expect HA-LoRA to focus only on encoding rich \textit{task-related} properties, while leaving the goal of selecting 
domain-invariant properties to domain-level purification.
For inference, we can merge HA-LoRA with the original weights of CLIP, eliminating extra
memory overhead.

\textbf{Head-aware LoRA \emph{vs.}~LoRA.}
Note that when jointly performing task-level and domain-level purification, head-ware LoRA is more beneficial to the domain-level purification than the convention LoRA.
It is because different to conventional LoRA, head-aware LoRA owns different learnable parameters across different heads, which avoids the interference between different heads.
Through such a head-interference elimination, the DIG may more independently and accordingly emphasize or restrain specific heads (see Section~\ref{ablation-halora} for empirical details).


\vspace{-2mm}
\subsection{Combine with prompt-learning methods}
\vspace{-2mm}
For classification with CLIP, we follow the general practice that comparing the similarity scores between the image feature and text features of difference classes.
To generate the text feature of a specific class, we need to conduct a text description which contains a prompt and the name of a class, and forward it through the text encoder of CLIP.
The quality of prompt also affects the generalization ability of CLIP.
Thus, many previous works, \emph{e.g.,} CoOp~\cite{coop}, CoCoOp~\cite{cocoop} and DPL~\cite{dpl} try to optimize the prompt to improve the generalization ability of CLIP.
In this paper, we focus on how to extract more generalizable image features from CLIP, which is orthogonal to the previous prompt-learning methods.
Thus, in practice, we may combine our method with previous prompt-learning method to maximize the generalization ability of CLIP across different domains, \emph{i.e.,} we may utilize previous prompt-learning methods to optimize the prompt and employ our attention head purification to optimize the image features.

\vspace{-2mm}
\section{Datasets and implementation details}
\vspace{-2mm}
\subsection{Datasets}
\vspace{-2mm}
For comparison, we evaluate our method on five representative datasets, including OfficeHome~\cite{officehome}, PACS~\cite{pacs}, VLCS~\cite{vlcs}, TerraIncognita~\cite{ti} and DomainNet~\cite{domainnet}. \textbf{OfficeHome} (OH) includes 4 domains 
with 15,588 examples from 65 classes. \textbf{VLCS} includes 4 domains 
with 10,729 examples from 5 classes. \textbf{PACS} includes 4 domains with 9,991 examples from 7 classes. \textbf{DomainNet} (DN) includes 6 domains 
with 586,575 examples from 345 classes. \textbf{TerraIncognita} (TI) includes 4 domains 
with 24788 examples from 10 classes.
Following~\cite{fourier,dann}, we adopt the typical leave-one-domain-out protocol for evaluation, \emph{i.e.,} 
each time we select one out of available domains as the target for testing and the remaining domains as sources for training.
The average accuracy across all target choices is reported.
\vspace{-2mm}
\subsection{Implementation Details}
\vspace{-2mm}
We use the CLIP pre-trained model with ViT-B/16 as the image encoder. 
We only tune the image encoder. The text encoder of CLIP is kept frozen throughout the training.
The batch size is set to 36. We use the AdamW as the optimizer with the cosine learning rate strategy for all datasets. We use a learning rate of $5\times10^{-5}$ for updating head-aware LoRA and $1\times10^{-3}$ for optimizing the head gates. For each run, we train the model for 40 epochs. We report the average result over three runs with different random seeds. We set the temperature $\tau$ to 0.01 which is the same as the pre-trained model. The $\alpha$ is set to 0.2 and kept the same across all the datasets. We purify the attention heads in all layers of the image encoder, and impose the MMD loss on each layer. For all the experiments, the rank of our head-aware LoRA is set to 8 for the last two layers and 2 for the rest layers.
\vspace{-2mm}
\section{Experimental Results} \label{sec: exp}
\vspace{-2mm}
We evaluate the proposed method for domain generalization classification task. We first conduct an extensive
ablation study to validate key components of our framework.
Second, we show that the proposed method can be combined
with prompt-learning techniques and obtain significant performance gain. This demonstrates that our
attention head purification technique is complementary to
a wide range of prompt-learning strategies and provides
additional benefits. Third, we show that our model performs favorably against previous state-of-the-art approaches. Finally, we visualize the attention maps of the overall model and specific heads.
\vspace{-2mm}
\subsection{Ablation Study}
\vspace{-2mm}
\noindent
\textbf{Task-level purification and domain-level purification cooperates to improve the generalization ability of CLIP.}
In Table~\ref{tab-component} (Left), we verify the effectiveness of task-level purification with head-aware LoRA (``HA-LoRA'') and domain-level purification with domain-invariant gating (``DIG''). We observe that both task-level purification and domain-level purification contribute to the performance improvement compared to the zero-shot baseline (the first line in Table~\ref{tab-component} (Left)). 
When combining both of them, we may further improve the domain generalization performance, obtaining 4.6\% gain on OfficeHome and 3.4\% gain on DomainNet compared to the zero-shot baseline.
All those results verify the effectiveness of our proposed task-level purification and domain-level purification operations. 

\noindent
\textbf{Head-aware LoRA eliminates interference between different heads, benefiting subsequent head selection in domain-level purification.}\label{ablation-halora}
In Table~\ref{tab-component} (Right), we investigate the effect of our proposed HA-LoRA compared with the conventional LoRA. We find that without performing domain-level purification (\emph{i.e.,} without using DIG), the result of HA-LoRA is only slightly better than that of the conventional LoRA (around 0.1\% gain). However, when jointly optimizing domain-invariant gates and LoRA/HA-loRA, HA-LoRA achieves a remarkable improvement compared to the conventional LoRA (more than 1\% gain). This is because the proposed head-aware LoRA can effectively eliminate the interference between different heads. 
Through interference elimination, the DIG may more independently and accordingly emphasize or restrain specific heads, rendering the domain-level purification more effective.  
\begin{table}
    \caption{\textbf{Left:} Effect of task/domain-level purification. \textbf{Right:} Head-aware LoRA \emph{vs.} original LoRA.}
  \begin{minipage}{0.5\linewidth}
  \centering
  \scalebox{0.85}{
  \begin{tabular}{cccc}
    \toprule
    HA-LoRA & DIG & OfficeHome&DomainNet \\
    \midrule
     &  & 82.4&57.7\\
    \checkmark & &85.0 &58.8 \\
    & \checkmark &83.6&58.2\\
    \checkmark & \checkmark &87.0&61.1 \\
    \bottomrule
  \end{tabular}}
  \end{minipage}%
  \begin{minipage}{0.5\linewidth}
  \centering
    \scalebox{0.8}{
      \begin{tabular}{ccccc}
    \toprule
    LoRA &HA-LoRA & DIG & OfficeHome &DomainNet \\
    \midrule
    \checkmark &  &w/o.& 84.8&58.7\\
     & \checkmark& w/o.&85.0\small\textit{(+0.2\%)}&58.8\small\textit{(+0.1\%)} \\
     \midrule
    \checkmark && w. &86.0&59.7\\
     & \checkmark & w. &87.0\small\textit{(+1.0\%)}&61.1\small\textit{(+1.4\%)} \\
    \bottomrule
  \end{tabular}
 }
  \end{minipage}
    \label{tab-component}
    \vspace{-5mm}
\end{table}
\vspace{-1mm}

\noindent
\textbf{Decoupling task-level purification and domain-level purification is beneficial.}\label{ablation-mmd}
{As discussed in Section~\ref{domainpuri}, we adopt MMD loss to encourage the head gates in DIG to update towards making the features invariant across domains. Technically, we may also adopt MMD loss to update HA-LoRA to encourage HA-LoRA to encode both task-related and domain-invariant properties. But we find this will harm the generalization performance. 
In Table~\ref{tab-mmd}, we compare the training without imposing any MMD loss (the first line), using MMD loss to update HA-LoRA (the second line), using MMD loss to update both the head gates and HA-LoRA (the third line) and our solution that uses MMD loss to update the head gates only (the last line). 
We observe that our solution obviously outperforms the one without any MMD loss, showing that MMD loss contributes to selecting/emphasizing the most domain-generalizable attention heads.
When applying MMD loss to update HA-LoRA, the performance decreases.
It is because applying MMD to HA-LoRA may encourage HA-LoRA to be both task-adapted and domain-invariant.
Such a coupled optimization is more difficult. 

\vspace{-1mm}
\noindent
\textbf{Joint purification training is the first choice.} 
In Table~\ref{tab-strate} (Left), we investigate the effect of different training strategies, including ours which jointly trains HA-LoRA and DIG, alternatively training DIG and HA-LoRA (denoted as alternative), training DIG first and then training HA-LoRA with fixed DIG (domain$\rightarrow$task), and training HA-LoRA first and then training DIG with fixed HA-LoRA (task$\rightarrow$domain). We observe that ours achieves the best results among different training strategies.

\begin{table}[h]
    \caption{Performance of using \(\mathcal{L}_\mathrm{MMD}\) to update different modules. 
    Results show that we may not obtain domain generalizable features directly through encouraging domain-invariant HA-LoRA.
    }
  \centering
  \label{tab-mmd}
  \scalebox{1}{
    \begin{tabular}{cccc}
    \toprule
    Modules that updated by \(\mathcal{L}_\mathrm{MMD}\)  & OfficeHome&DomainNet&PACS \\
    \midrule
    Neither   &  86.2 &60.5 &96.9  \\
    HA-LoRA  & 85.8 &58.6 & 96.1 \\
    HA-LoRA + DIG &  86.0 &59.9&96.7 \\
    \rowcolor[gray]{.9}
    DIG &  87.0 &61.1&98.1 \\
    \bottomrule
  \end{tabular}}
  \vspace{-3mm}
\end{table}




\begin{table}[h!]
    \caption{\textbf{Left:} Effect of different training strategies.  \textbf{Right:} Sensitivity to the ratio $\alpha$ of MMD loss term. The experiments are conducted on OfficeHome. The trends are similar for the other datasets.}
  \begin{minipage}{0.5\linewidth}
  \centering
  \scalebox{1}{
   \begin{tabular}{lc}
    \toprule
    Method     & OfficeHome \\
    \midrule
    jointly &  87.0  \\
    alternative & 86.7\\
    two-stage(Task $\rightarrow$ Domain) &  85.8  \\
    two-stage(Domain $\rightarrow$ Task) &  85.1  \\
    \bottomrule
\end{tabular}}
  \end{minipage}%
  \begin{minipage}{0.5\linewidth}
  \centering
    \scalebox{1}{
  \begin{tabular}{cc}
    \toprule
    Values of $\alpha$ & OfficeHome \\
    \midrule
    0.0 &  86.2  \\
    0.1 & 86.7\\
    0.2 & 87.0   \\
    0.3 & 86.9   \\
    0.5& 86.7\\
    \bottomrule
\end{tabular}
 }
  \end{minipage}
    \label{tab-strate}
\vspace{-3mm}
\end{table}

\textbf{Sensitivity to the ratio \(\alpha\) of MMD loss term.}
In Table~\ref{tab-strate} (Right), we evaluate how the performance changes as $\alpha$ increases.
We observe as $\alpha$ increases, the accuracy firstly increases and then decreases, exhibiting a typical bell curve. 
This phenomenon shows the regularization effect of MMD loss term.
Besides, within a vast range of $\alpha$, the accuracy only slightly fluctuates. 
Note that the trends are similar across all the datasets. 
The results show that our method is relatively robust to the choices of $\alpha$.
\vspace{-2mm}
\subsection{Improvement beyond prompt-learning methods} \label{pt}
\vspace{-2mm}
In Table~\ref{table-combine}, we combine our method with representative prompt-learning DG methods. 
Besides, we also report numbers obtained by our method with manually designed prompt ``A photo of a'' (same as zero-shot CLIP baseline).
We observe that even without any prompt optimization, our method achieves remarkable improvement compared to zero-shot CLIP baseline (around 8\% gain).
Previous prompt-learning DG methods can be roughly categorized into two groups. One group, \emph{e.g.,} DUPRG and PromptStyler, only utilizes text features to optimize the prompt. The other group, \emph{e.g.,} CoOp, CoCoOp, DPL and STYLIP, utilizes the alignment between image features and text features to optimize the learnable prompt.
Thus, for different groups, we combine attention head purification with prompt-learning in different ways.
For the first group, we sequentially perform prompt optimization and attention head purification, \emph{i.e.,}, we firstly obtain optimized prompt with prompt-learning method and then utilize the optimized prompt in attention head purification learning. 
For the second group, we jointly learn attention head purification and optimize the prompt.
As shown in Table~\ref{table-combine}, combining with attention head purification can consistently improve the generalization performance of CLIP beyond the prompt-learning methods, \emph{e.g.,} combining attention head purification with PromptStyler yields the best result (79.0\% average accuracy) with 5.5\% improvement beyond PromptStyler.

\begin{table}[ht]
  \caption{Improvement beyond prompt-learning methods. \textsuperscript{\dag} indicates that the number is reproduced by us since the number is not provided in the original paper. Others are cited from the original paper.}
  \label{table-combine}
  \centering

  \begin{tabular}{ccccccc}
    \toprule
    Method     & OH  &VLCS&PACS& DN&TI& Avg. \\
    \midrule
    Zero-Shot&82.4&81.7&96.1&56.6&33.8&70.1\\
    \vspace{1mm}
    \ \ \ \ \textit{+ours} &87.0 &85.1&98.1&61.1&59.7&78.2 \small \textit{(+8.1\%)} \\
    CoOp\textsuperscript{\dag}~\cite{coop}&83.0&80.8&96.4&59.5&46.8&73.6\\
    \vspace{1mm}
    \ \ \ \ \textit{+ours} &87.3 &85.3&98.2&61.0&59.9&78.3 \small \textit{(+4.7\%)} \\
    
    CoCoOp\textsuperscript{\dag}~\cite{cocoop}&83.4&80.3&96.7&59.4&45.3&73.2\\
    \vspace{1mm}
    \ \ \ \ \textit{+ours} &87.4 &84.8&98.4&61.3&58.8&78.2 \small \textit{(+5.0\%)} \\
    
    DPL~\cite{dpl}&84.2&84.3&97.3&56.7&52.6\textsuperscript{\dag}&75.0\\
    \vspace{1mm}
    \ \ \ \ \textit{+ours} &87.2 &85.1&98.0&61.4&60.6 &78.5 \small \textit{(+3.5\%)} \\
    
    DUPRG~\cite{duprg}&83.6&83.9&97.1&59.6&42.0&73.2\\
    \vspace{1mm}
    \ \ \ \ \textit{+ours}&87.0&85.4&98.2&61.5&60.1&78.4\small \textit{(+5.2\%)}\\
    STYLIP\textsuperscript{\dag}~\cite{stylip}&84.1&84.8&96.8&59.9&57.4&76.6\\   
    \vspace{1mm}
    \ \ \ \ \textit{+ours}&87.5&85.3&98.5&62.1&59.9&78.7\small \textit{(+2.1\%)}\\
 
    PromptStyler~\cite{promptstyler}&83.6&82.9&97.2&59.4&44.2\textsuperscript{\dag}&73.5\\
    \ \ \ \ \textit{+ours}&87.7&86.1&98.3&62.0&60.6&79.0\small \textit{(+5.5\%)}\\

    \bottomrule
  \end{tabular}
  \vspace{-3mm}
\end{table}

\subsection{Comparison with Previous State-of-the-Arts}
\vspace{-2mm}
In Table~\ref{tab: main-compare}, we compare our solution with previous state-of-the-art CLIP-based domain generalization methods.
Besides, we also compare our method to baselines including zero-shot CLIP (``zero-shot''), linear probing of CLIP (``Linear-Probe''), and standard full fine-tuning of CLIP with all source domains (``ERM-FFT''). 
We report the accuracy for each dataset and the average accuracy across all the datasets in Table~\ref{tab: main-compare} for comparison.
\textsuperscript{\dag} indicates that the number is reproduced by us since it is not provided in the original paper. 
Other numbers are cited from the original paper.
For our method, we report numbers obtained by attention head purification combined with prompt learning (``Ours'').

\begin{table}[ht]
  \caption{Comparison with previous state-of-the-arts. 
  \textsuperscript{\dag} indicates that the number is reproduced by us as it is not reported by the original paper. Others are cited from the original paper.}
  \label{sample-table}
  \centering
  
  \begin{tabular}{lcccccc}
    \toprule
    Method
    & OH  &VLCS&PACS& DN&TI& Avg. \\
    \midrule
    Zero-Shot 
    & 82.4 &81.7&96.1&56.6 &33.8&70.1  \\
    Linear-Probe\textsuperscript{\dag}
    &79.3&77.5&94.9&48.2&44.6&68.9\\
    ERM-FFT\textsuperscript{\dag}
    &80.0&79.1&91.4&53.9&44.1&69.7\\
    CoOp\textsuperscript{\dag}~\cite{coop}
    &83.0&80.8&96.4&59.5&46.8&73.6\\
    CoCoOp\textsuperscript{\dag}~\cite{cocoop}
    &83.4&80.3&96.7&59.4&45.3&73.2\\
    MaPLe\textsuperscript{\dag}~\cite{maple}
    &83.4&82.2&96.5&59.5&50.2&74.4\\
    VPT\textsuperscript{\dag}~\cite{vpt}
    &83.2&82.0&96.9&58.5&46.7&73.6\\

    DPL~\cite{dpl}
    &84.2&84.3&97.3&56.7&52.6\textsuperscript{\dag}&75.0\\
    DUPRG~\cite{duprg}
    &83.6&83.9&97.1&59.6&42.0&73.2\\
    PromptStyler~\cite{promptstyler}
    &83.6&82.9&97.2&59.4&44.2\textsuperscript{\dag}&73.5\\
    STYLIP~\cite{stylip}
    &84.6&\textbf{86.9}&98.1&62.0&57.4\textsuperscript{\dag}&77.8\\
    DSPL~\cite{dspl}
    &86.1&86.4&97.5&62.1&57.1&77.8\\
    SPG~\cite{spg}
    &83.6&82.4&97.0&60.1&50.2&74.7\\
    
    VL2V-SD~\cite{vl2v-sd}
    & 85.4 &82.7&95.7&58.7&41.2&72.7\\  
    CLIP-LoRA\textsuperscript{\dag}~\cite{cliplora}
    &83.9&83.1&97.1&58.4&55.7&75.6\\
    GESTUR~\cite{gestur}
    &84.2&82.8&96.0&58.9&55.7&75.5\\
    MIRO~\cite{5}
    & 82.5 &82.2&95.6&54.0&54.3&73.7\\
    CLIPood~\cite{clipood}
    & 87.0 &85.0&97.3&\textbf{63.5}&60.4&78.6\\
 CLIPood\textsuperscript{\dag}~\cite{clipood}&85.3&83.6&97.3&59.5&58.7&76.9\\
    \rowcolor[gray]{.9}Ours&\textbf{87.7}&86.1&\textbf{98.3}&62.0&\textbf{60.6}&\textbf{79.0}\\
    \bottomrule
    \label{tab: main-compare}
    \end{tabular}
  \vspace{-5mm}
\end{table}

From Table~\ref{tab: main-compare}, we find that CLIP's zero-shot result serves as a strong baseline. 
Directly linear probing or full fine-tuning CLIP's image encoder even results in worse accuracy, demonstrating that it is non-trivial to harness CLIP for domain generalization tasks.
For methods that fine-tune the image encoder, existing competitors, including GESTUR, MIRO, and CLIPood, mainly focus on avoiding knowledge forgetting of CLIP during task adaptation. Our method outperforms these methods, \emph{e.g.,} 
outperforming MIRO by more than 5\%,
indicating that our way of performing attention head purification is more effective on improving the CLIP's domain generalization performance.
Overall, compared with various CLIP-based generalization methods, our method achieves the best result.

\begin{figure}[h!]
    \centering
    \includegraphics[width=1\textwidth]{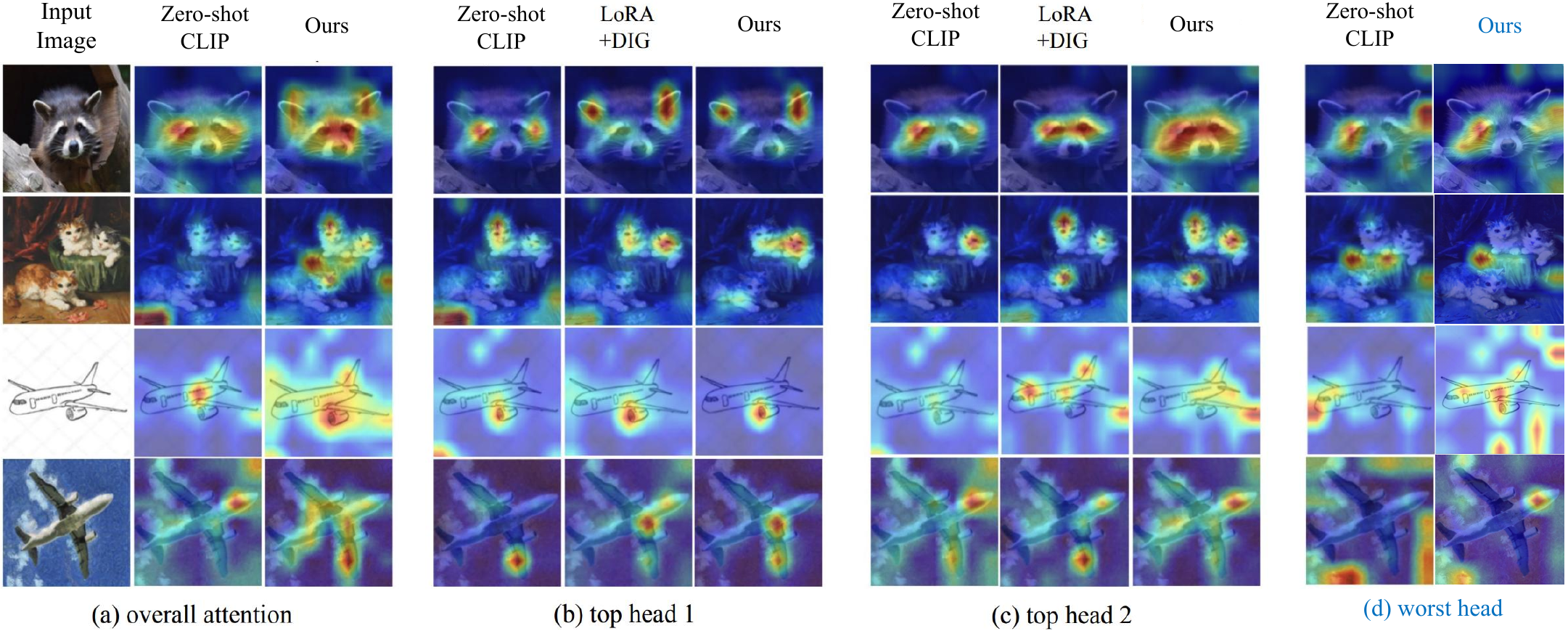}
    \caption{Attention maps for target samples in DomainNet.} 
    \label{fig-atten}
    \vspace{-6mm}
\end{figure}

\subsection{Visualization}
\vspace{-1mm}
We visualize the overall attention maps~\cite{attentionmap} before and after attention head purification in Figure~\ref{fig-atten}(a).
We observe that the zero-shot model owns a sparser attention and pays more attention to the background which may not be generalizable across domains. In contrast, ours attends to the most discriminative and generalizable properties of the object for classification.
We further visualize the attention maps generated by specific attention heads. We show the attention maps of the top two heads (Figure \ref{fig-atten}(b)(c)) and the last head (Figure \ref{fig-atten}(d)) ranked by the learned head gates. The top two heads exhibit different preferences. Generally, top head 1 tends to capture the most discriminative part of each object, for example, the ear of a raccoon (first row) and the engine of a plane (third row), while top head 2 tends to focus on the overall region of an object. Besides, compared with purifying heads using conventional LoRA with DIG (middle column in Figure~\ref{fig-atten} (b)(c)), thanks to the interference elimination of HA-LoRA, our method can better highlight the perceptual tendency of a specific head. 
For the worst head, as shown in Figure \ref{fig-atten}(d), background areas receive high attention. By assigning less weight to such heads in our method, their influence is suppressed.

\vspace{-2mm}
\section{Conclusion}
\vspace{-3mm}
In this paper, we propose a simple yet effective method to harness CLIP for domain generalization, \emph{i.e.,} attention head purification. Specifically, we perform attention head purification from two perspectives including task-level purification and domain-level purification. For task-level purification, we design head-aware LoRA to make each head specifically adapted to the downstream tasks. For domain-level purification, we adopt a domain-invariant gating strategy to encourage the model to select/emphasize the most generalizable attention heads. During training, we jointly perform the task-level purification and the domain-level purification. Experiments on five representative domain generalization benchmarks demonstrate the superiority of our proposed method. 


\appendix
\section{Appendix}
\subsection{Attention head evaluation strategy mentioned in Figure~\ref{fig-overview}}
In Figure~\ref{fig-overview}(a), we evaluate the importance of attention heads for domain generalization tasks with different strategies including ``randomly drop'', ``manually drop'', ``drop by cross-validation'', and ``adapt \& drop''. We introduce the implementation details of each strategy as follows:

\textbf{Randomly drop}: we randomly sample a certain number of attention heads and drop them for inference. We repeat the random sampling three times and report the average results.

\textbf{Manually drop}: we manually evaluate the generalization performance of each head based on the text describing the properties of each head provided in~\cite{interpret} and then drop the least generalizable heads for inference.

\textbf{Drop by cross-validation}: 
Each time we select one domain out of available domains as the target domain for evaluation. We use the remaining domains to train the model to learn the importance of each head for domain generalization task. Specifically, we adopt $\mathcal{L}_{cls}$ to train a learnable Bernoulli gate for each head representing the probability to remain the head, with the help of Gumbel-Softmax trick~\cite{gumble}. During inference, we drop heads with probability from lowest to highest. 
The CLIP is frozen during training.
We report the average accuracy over different target domains.

\textbf{Adapt \& drop}: following the above cross-validation setup, we perform attention head adaptation with LoRA and learn the gates simultaneously.

\begin{figure}[h!]
    \centering
    \includegraphics[width=1\textwidth]{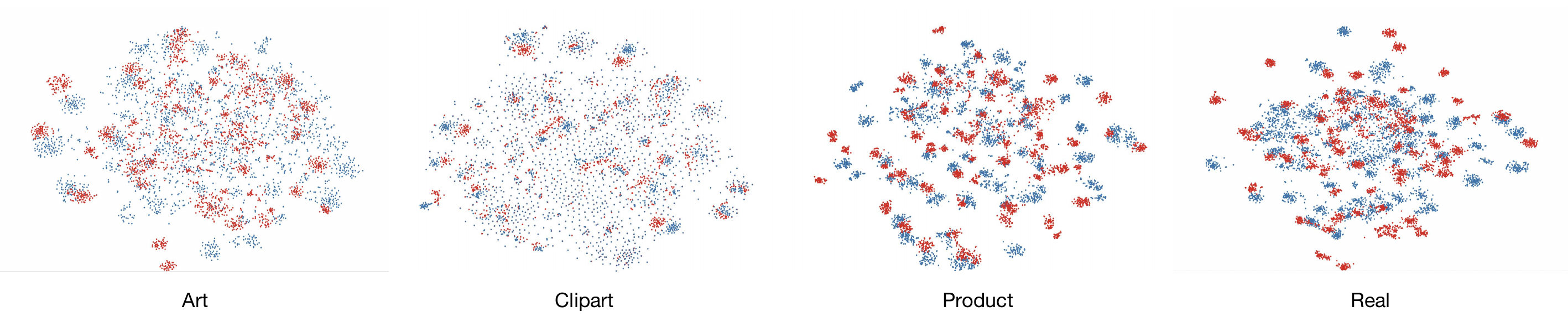}
    \caption{t-SNE visualization of features from target domain on OfficeHome. We compare the visualization results of the original feature (black dot) and our method (red dot).} 
    \label{fig-tsne}
\end{figure}
\begin{table}[!h]
 \centering
    \caption{\textbf{Left:} Effect of soft gating in domain-level purification. \textbf{Right:} Training and inference time on DomainNet. Experiments were conducted on a single RTX4090 24Gb with the original code provided by the authors.}
  \begin{minipage}{0.5\linewidth}
  \centering
  \scalebox{1}{
  \begin{tabular}{cccc}
    \toprule
    Gating Strategy & PACS & OfficeHome&DomainNet \\
    \midrule
    binary mask & 96.9&85.4 &57.7 \\
     soft gating&  98.1& 87.0&61.1\\
    \bottomrule
  \end{tabular}}
  \end{minipage}%
  \begin{minipage}{0.5\linewidth}
  \centering
    \scalebox{1}{
     \begin{tabular}{lcc}
    \toprule
    Method & Training time&\textcolor{black}{Inference time} \\
    \midrule
    CoOp &  2h 57min.&39s  \\
    DPL & 2h&41s\\
    MIRO & 5h 24min.&39s   \\
    CLIPood & 4h 46min.&39s   \\
    Ours& 1h 30min.&39s\\
    \bottomrule
\end{tabular}}
  \end{minipage}
    \label{tab-gum}
\end{table}

\subsection{Performing DIG using binary mask instead of soft gating}
With the help of Gumbel-Softmax trick, we can generate binary mask for attention heads to fully retain or remove a specific head. In Table~\ref{tab-gum} (Left), we provide the results of replacing soft gating with binary mask in DIG. We find that soft gating yields better performance. 

\subsection{Visualization with t-SNE}
We present the t-SNE visualization of the feature distribution on OfficeHome in Figure~\ref{fig-tsne}. The black dots denote CLIP's original visual feature and the red dots denote visual features generated by our method. For the original visual features, feature distribution is more dispersed 
especially for the domain Clipart. Nevertheless, benefiting from the attention head purification, the features of ours are more compact and the distribution is more concentrated, which is in line with the superior domain generalization performance of our method.

\subsection{Computational efficiency}
Table~\ref{tab-gum} (Right) compares the training time of the leading prompt-learning methods (CoOp~\cite{coop} and DPL~\cite{dpl}) and fine-tuning methods (MIRO~\cite{5} and CLIPood~\cite{clipood}). Our method achieves better performance
with shorter training time. 
\textcolor{black}{We evaluate the inference time for each method on the ``Real'' domain of DomainNet dataset, with batch size set to 128. We observe that our method doesn't introduce additional inference time compared to previous works.}

\subsection{Effect of Fine-tuning the text encoder}
\textcolor{black}{
The text features with specifically designed prompt in nature contain rare domain-specific information. Additionally, co-adapting the image encoder and text encoder on limited data can lead to overfitting, potentially disrupting the alignment between image and text features. As a result, we freeze the text encoder. As shown in Table~\ref{table-textencoder}, co-adapting the image and text encoders results in notable performance degradation.}

\begin{table}[ht]
  \caption{\textcolor{black}{Effect of fine-tuning CLIP's text encoder. We compare the results of co-adapting image encoder and text encoder with those of ours which freezes the text encoder.}}
  \label{table-textencoder}
  \centering
  \scalebox{1}{
  \begin{tabular}{lcc}
    \toprule
    Method     & OH  &PACS \\
    \midrule
    Ours w. fine-tuning text encoder&84.7&96.5\\
    \rowcolor[gray]{.9}
    Ours w.o. fine-tuning text encoder&87.0&98.1\\
    \bottomrule
  \end{tabular}}
\end{table}

\subsection{Visualizations with overall attention}
\textcolor{black}{
In Figure~\ref{fig-overallattention}, we compare the overall attention maps of Head-Aware LoRA with DIG (our method), conventional LoRA with DIG, and the regularization-based method CLIPood. Our method generates better attention, enabling a more accurate and comprehensive perception of discriminative parts of the target object.}
\begin{figure}[h!]
    \centering
    \includegraphics[width=1\textwidth]{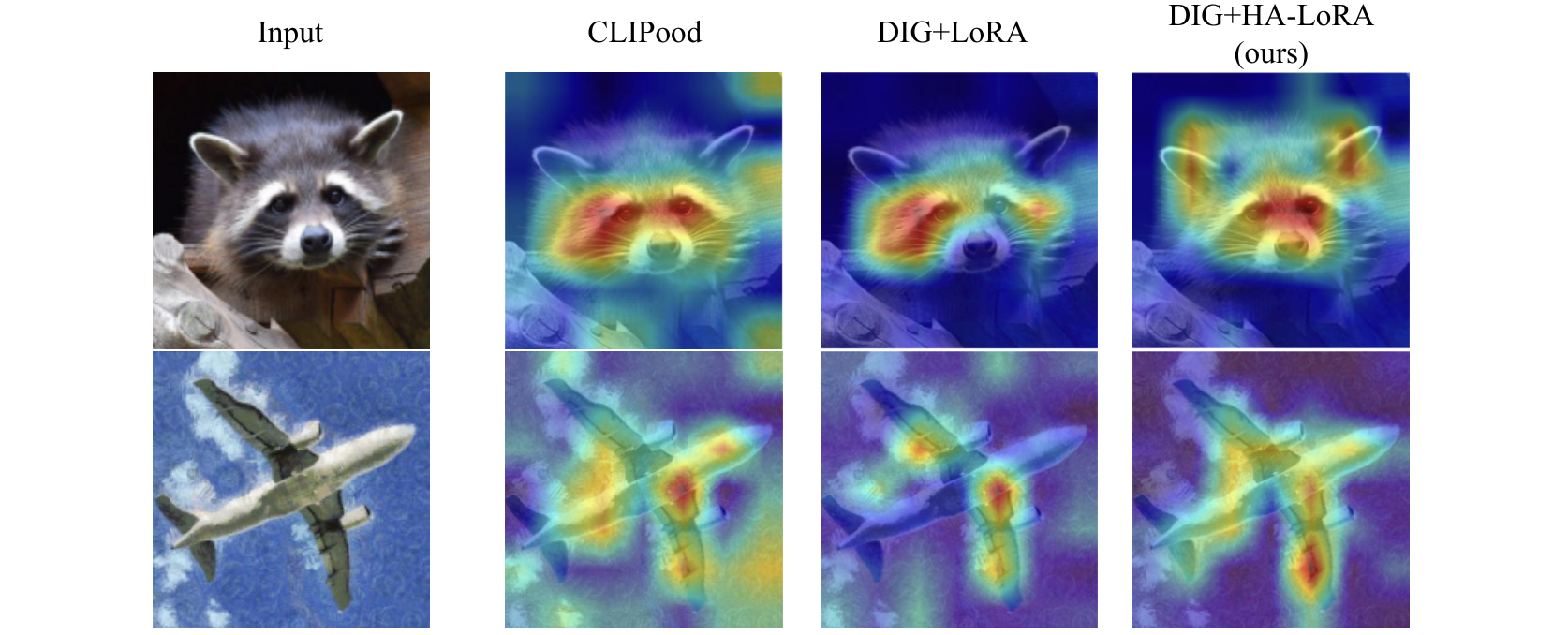}
    \caption{\textcolor{black}{Comparison of the overall attention. Images are sampled from the DomainNet dataset. The ``Real'' is selected as the target domain while others are selected as source domains to train the CLIP.}}
    \label{fig-overallattention}
\end{figure}

\subsection{Visualization with learned gating weights of DIG}
\textcolor{black}{In Figure~\ref{fig-weight}, we visualize the learned gating weights of DIG. The observation is that the weight distribution is non-uniform, and there exists an apparent gap between the largest weight and the smallest weight. The results demonstrate that with DIG, some heads are relatively emphasized and some heads are relatively suppressed.} 
\begin{figure}[h!]
    \centering
    \includegraphics[width=1\textwidth]{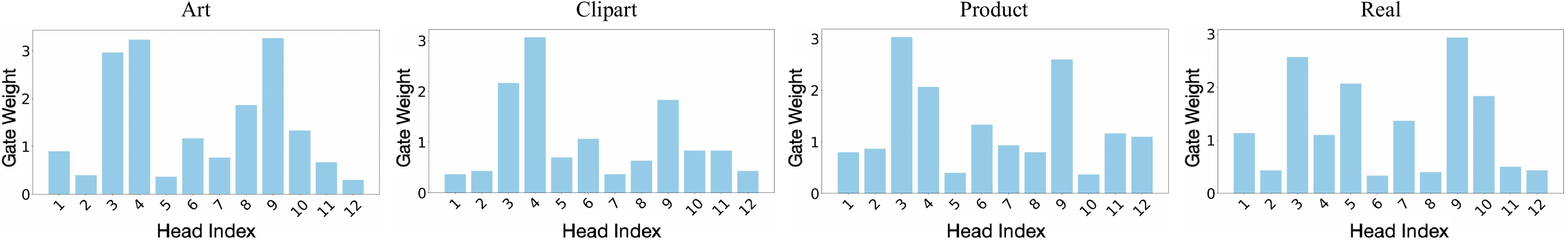}
    \caption{\textcolor{black}{Distribution of the learned gating weights of DIG. Models are trained on OfficeHome. The domain name refers to the target domain in each case. The gate weights are not within 0-1 since we multiply the gates after Softmax operation \(\hat{g}\) by the number of heads \(\gamma\), as shown in Equation~(\ref{eq-gate}).}}
    \label{fig-weight}
\end{figure}

\subsection{Effectiveness of DIG in removing domain-specific information.}
\textcolor{black}{
In Figure~\ref{fig-dig}, we provide attention to show the effectiveness of DIG in removing domain-specific information. The attention of different heads is ranked by the weights of DIG. For illustration purposes, we visualize the overall attention by aggregating the attention from all the heads.
Purely using HA-LoRA, the weights of different head attentions are equal to 1 for aggregating, 
while with DIG, the weights of different heads are different.
From DIG ranking, we observe that the attention map which focuses on the discriminative parts of target object owns a larger gate weight while the attention map which focuses on the object-irrelevant background area owns a smaller gate weight. 
As a result, 
we observe that with the application of DIG, the domain-specific components (\emph{e.g.}, background regions) in the overall attention are effectively suppressed, allowing better focus on the target object (\emph{e.g.,} the monkey).}
\begin{figure}[h!]
    \centering
    \includegraphics[width=1\textwidth]{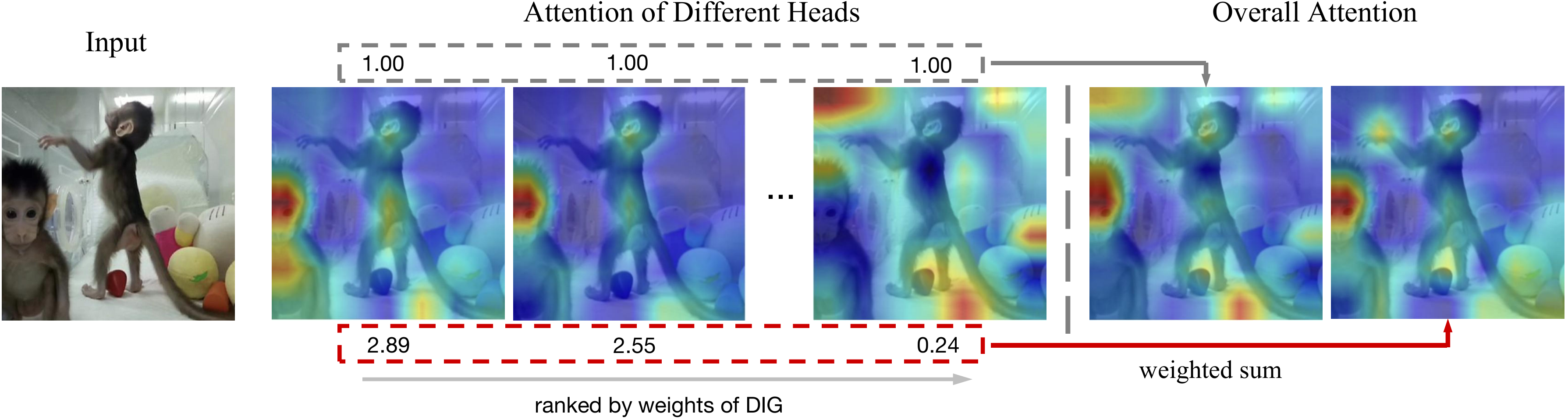}
    \caption{\textcolor{black}{Effectiveness of DIG in removing domain-specific information. For illustration purpose, the overall attention is computed as a weighted sum of attention from different heads, using either DIG weights (as shown in red color) or uniform weights (\emph{i.e.,} without DIG, as shown in gray color).} }
    \label{fig-dig}
\end{figure}

\bibliographystyle{unsrt}  
\bibliography{references}

\begin{thebibliography}{10}

\bibitem{1}
Fabio~M Carlucci, Antonio D'Innocente, Silvia Bucci, Barbara Caputo, and Tatiana Tommasi.
\newblock Domain generalization by solving jigsaw puzzles.
\newblock In {\em Proceedings of the IEEE/CVF Conference on Computer Vision and Pattern Recognition}, pages 2229--2238, 2019.

\bibitem{5}
Junbum Cha, Kyungjae Lee, Sungrae Park, and Sanghyuk Chun.
\newblock Domain generalization by mutual-information regularization with pre-trained models.
\newblock In {\em European conference on computer vision}, pages 440--457. Springer, 2022.

\bibitem{12}
Ishaan Gulrajani and David Lopez-Paz.
\newblock In search of lost domain generalization.
\newblock {\em arXiv preprint arXiv:2007.01434}, 2020.

\bibitem{13}
Donghyun Kim, Kaihong Wang, Stan Sclaroff, and Kate Saenko.
\newblock A broad study of pre-training for domain generalization and adaptation.
\newblock In {\em European Conference on Computer Vision}, pages 621--638. Springer, 2022.

\bibitem{15}
Da~Li, Yongxin Yang, Yi-Zhe Song, and Timothy Hospedales.
\newblock Learning to generalize: Meta-learning for domain generalization.
\newblock In {\em Proceedings of the AAAI conference on artificial intelligence}, volume~32, 2018.

\bibitem{18}
Haoliang Li, Sinno~Jialin Pan, Shiqi Wang, and Alex~C Kot.
\newblock Domain generalization with adversarial feature learning.
\newblock In {\em Proceedings of the IEEE conference on computer vision and pattern recognition}, pages 5400--5409, 2018.

\bibitem{mix-style}
Kaiyang Zhou, Yongxin Yang, Yu~Qiao, and Tao Xiang.
\newblock Domain generalization with mixstyle.
\newblock {\em arXiv preprint arXiv:2104.02008}, 2021.

\bibitem{3}
Jindong Wang, Cuiling Lan, Chang Liu, Yidong Ouyang, Tao Qin, Wang Lu, Yiqiang Chen, Wenjun Zeng, and S~Yu Philip.
\newblock Generalizing to unseen domains: A survey on domain generalization.
\newblock {\em IEEE transactions on knowledge and data engineering}, 35(8):8052--8072, 2022.

\bibitem{4}
Yunpei Jia, Jie Zhang, Shiguang Shan, and Xilin Chen.
\newblock Single-side domain generalization for face anti-spoofing.
\newblock In {\em Proceedings of the IEEE/CVF Conference on Computer Vision and Pattern Recognition}, pages 8484--8493, 2020.

\bibitem{kl1}
Haoliang Li, YuFei Wang, Renjie Wan, Shiqi Wang, Tie-Qiang Li, and Alex Kot.
\newblock Domain generalization for medical imaging classification with linear-dependency regularization.
\newblock {\em Advances in neural information processing systems}, 33:3118--3129, 2020.

\bibitem{adv2}
Ya~Li, Xinmei Tian, Mingming Gong, Yajing Liu, Tongliang Liu, Kun Zhang, and Dacheng Tao.
\newblock Deep domain generalization via conditional invariant adversarial networks.
\newblock In {\em Proceedings of the European conference on computer vision (ECCV)}, pages 624--639, 2018.

\bibitem{adv3}
Rui Shao, Xiangyuan Lan, Jiawei Li, and Pong~C Yuen.
\newblock Multi-adversarial discriminative deep domain generalization for face presentation attack detection.
\newblock In {\em Proceedings of the IEEE/CVF conference on computer vision and pattern recognition}, pages 10023--10031, 2019.

\bibitem{kl2}
Ziqi Wang, Marco Loog, and Jan Van~Gemert.
\newblock Respecting domain relations: Hypothesis invariance for domain generalization.
\newblock In {\em 2020 25th International Conference on Pattern Recognition (ICPR)}, pages 9756--9763. IEEE, 2021.

\bibitem{meta1}
Swami Sankaranarayanan and Yogesh Balaji.
\newblock Meta learning for domain generalization.
\newblock In {\em Meta Learning With Medical Imaging and Health Informatics Applications}, pages 75--86. Elsevier, 2023.

\bibitem{metareg}
Yogesh Balaji, Swami Sankaranarayanan, and Rama Chellappa.
\newblock Metareg: Towards domain generalization using meta-regularization.
\newblock {\em Advances in neural information processing systems}, 31, 2018.

\bibitem{meta2}
Yogesh Balaji, Swami Sankaranarayanan, and Rama Chellappa.
\newblock Metareg: Towards domain generalization using meta-regularization.
\newblock {\em Advances in neural information processing systems}, 31, 2018.

\bibitem{advaug}
Riccardo Volpi, Hongseok Namkoong, Ozan Sener, John~C Duchi, Vittorio Murino, and Silvio Savarese.
\newblock Generalizing to unseen domains via adversarial data augmentation.
\newblock {\em Advances in neural information processing systems}, 31, 2018.

\bibitem{aug1}
Fengchun Qiao, Long Zhao, and Xi~Peng.
\newblock Learning to learn single domain generalization.
\newblock In {\em Proceedings of the IEEE/CVF conference on computer vision and pattern recognition}, pages 12556--12565, 2020.

\bibitem{aug2}
Kaiyang Zhou, Yongxin Yang, Timothy Hospedales, and Tao Xiang.
\newblock Learning to generate novel domains for domain generalization.
\newblock In {\em Computer Vision--ECCV 2020: 16th European Conference, Glasgow, UK, August 23--28, 2020, Proceedings, Part XVI 16}, pages 561--578. Springer, 2020.

\bibitem{imgnet}
Minyoung Huh, Pulkit Agrawal, and Alexei~A Efros.
\newblock What makes imagenet good for transfer learning?
\newblock {\em arXiv preprint arXiv:1608.08614}, 2016.

\bibitem{itpairs1}
Alec Radford, Jong~Wook Kim, Chris Hallacy, Aditya Ramesh, Gabriel Goh, Sandhini Agarwal, Girish Sastry, Amanda Askell, Pamela Mishkin, Jack Clark, et~al.
\newblock Learning transferable visual models from natural language supervision.
\newblock In {\em International conference on machine learning}, pages 8748--8763. PMLR, 2021.

\bibitem{itpairs2}
Chao Jia, Yinfei Yang, Ye~Xia, Yi-Ting Chen, Zarana Parekh, Hieu Pham, Quoc Le, Yun-Hsuan Sung, Zhen Li, and Tom Duerig.
\newblock Scaling up visual and vision-language representation learning with noisy text supervision.
\newblock In {\em International conference on machine learning}, pages 4904--4916. PMLR, 2021.

\bibitem{cost1}
Hieu Pham, Zihang Dai, Golnaz Ghiasi, Kenji Kawaguchi, Hanxiao Liu, Adams~Wei Yu, Jiahui Yu, Yi-Ting Chen, Minh-Thang Luong, Yonghui Wu, et~al.
\newblock Combined scaling for zero-shot transfer learning.
\newblock {\em Neurocomputing}, 555:126658, 2023.

\bibitem{cost2}
Mitchell Wortsman, Gabriel Ilharco, Jong~Wook Kim, Mike Li, Simon Kornblith, Rebecca Roelofs, Raphael~Gontijo Lopes, Hannaneh Hajishirzi, Ali Farhadi, Hongseok Namkoong, et~al.
\newblock Robust fine-tuning of zero-shot models.
\newblock In {\em Proceedings of the IEEE/CVF conference on computer vision and pattern recognition}, pages 7959--7971, 2022.

\bibitem{clipood}
Yang Shu, Xingzhuo Guo, Jialong Wu, Ximei Wang, Jianmin Wang, and Mingsheng Long.
\newblock Clipood: Generalizing clip to out-of-distributions.
\newblock In {\em International Conference on Machine Learning}, pages 31716--31731. PMLR, 2023.

\bibitem{attentionmap}
Hila Chefer, Shir Gur, and Lior Wolf.
\newblock Generic attention-model explainability for interpreting bi-modal and encoder-decoder transformers.
\newblock In {\em Proceedings of the IEEE/CVF International Conference on Computer Vision}, pages 397--406, 2021.

\bibitem{officehome}
Hemanth Venkateswara, Jose Eusebio, Shayok Chakraborty, and Sethuraman Panchanathan.
\newblock Deep hashing network for unsupervised domain adaptation.
\newblock In {\em Proceedings of the IEEE conference on computer vision and pattern recognition}, pages 5018--5027, 2017.

\bibitem{interpret}
Yossi Gandelsman, Alexei~A Efros, and Jacob Steinhardt.
\newblock Interpreting clip's image representation via text-based decomposition.
\newblock {\em arXiv preprint arXiv:2310.05916}, 2023.

\bibitem{domainnet}
Xingchao Peng, Qinxun Bai, Xide Xia, Zijun Huang, Kate Saenko, and Bo~Wang.
\newblock Moment matching for multi-source domain adaptation.
\newblock In {\em Proceedings of the IEEE/CVF international conference on computer vision}, pages 1406--1415, 2019.

\bibitem{pacs}
Da~Li, Yongxin Yang, Yi-Zhe Song, and Timothy~M Hospedales.
\newblock Deeper, broader and artier domain generalization.
\newblock In {\em Proceedings of the IEEE international conference on computer vision}, pages 5542--5550, 2017.

\bibitem{vlcs}
Antonio Torralba and Alexei~A Efros.
\newblock Unbiased look at dataset bias.
\newblock In {\em CVPR 2011}, pages 1521--1528. IEEE, 2011.

\bibitem{ti}
Sara Beery, Grant Van~Horn, and Pietro Perona.
\newblock Recognition in terra incognita.
\newblock In {\em Proceedings of the European conference on computer vision (ECCV)}, pages 456--473, 2018.

\bibitem{devise}
Andrea Frome, Greg~S Corrado, Jon Shlens, Samy Bengio, Jeff Dean, Marc'Aurelio Ranzato, and Tomas Mikolov.
\newblock Devise: A deep visual-semantic embedding model.
\newblock {\em Advances in neural information processing systems}, 26, 2013.

\bibitem{socher2013zero}
Richard Socher, Milind Ganjoo, Christopher~D Manning, and Andrew Ng.
\newblock Zero-shot learning through cross-modal transfer.
\newblock {\em Advances in neural information processing systems}, 26, 2013.

\bibitem{write}
Mohamed Elhoseiny, Babak Saleh, and Ahmed Elgammal.
\newblock Write a classifier: Zero-shot learning using purely textual descriptions.
\newblock In {\em Proceedings of the IEEE International Conference on Computer Vision}, pages 2584--2591, 2013.

\bibitem{attention}
Ashish Vaswani, Noam Shazeer, Niki Parmar, Jakob Uszkoreit, Llion Jones, Aidan~N Gomez, {\L}ukasz Kaiser, and Illia Polosukhin.
\newblock Attention is all you need.
\newblock {\em Advances in neural information processing systems}, 30, 2017.

\bibitem{lxmert}
Hao Tan and Mohit Bansal.
\newblock Lxmert: Learning cross-modality encoder representations from transformers.
\newblock {\em arXiv preprint arXiv:1908.07490}, 2019.

\bibitem{vl}
Weijie Su, Xizhou Zhu, Yue Cao, Bin Li, Lewei Lu, Furu Wei, and Jifeng Dai.
\newblock Vl-bert: Pre-training of generic visual-linguistic representations.
\newblock {\em arXiv preprint arXiv:1908.08530}, 2019.

\bibitem{kim2021vilt}
Wonjae Kim, Bokyung Son, and Ildoo Kim.
\newblock Vilt: Vision-and-language transformer without convolution or region supervision.
\newblock In {\em International conference on machine learning}, pages 5583--5594. PMLR, 2021.

\bibitem{li2021supervision}
Yangguang Li, Feng Liang, Lichen Zhao, Yufeng Cui, Wanli Ouyang, Jing Shao, Fengwei Yu, and Junjie Yan.
\newblock Supervision exists everywhere: A data efficient contrastive language-image pre-training paradigm.
\newblock {\em arXiv preprint arXiv:2110.05208}, 2021.

\bibitem{slip}
Norman Mu, Alexander Kirillov, David Wagner, and Saining Xie.
\newblock Slip: Self-supervision meets language-image pre-training.
\newblock In {\em European conference on computer vision}, pages 529--544. Springer, 2022.

\bibitem{zhai2022lit}
Xiaohua Zhai, Xiao Wang, Basil Mustafa, Andreas Steiner, Daniel Keysers, Alexander Kolesnikov, and Lucas Beyer.
\newblock Lit: Zero-shot transfer with locked-image text tuning.
\newblock In {\em Proceedings of the IEEE/CVF Conference on Computer Vision and Pattern Recognition}, pages 18123--18133, 2022.

\bibitem{cyclip}
Shashank Goel, Hritik Bansal, Sumit Bhatia, Ryan Rossi, Vishwa Vinay, and Aditya Grover.
\newblock Cyclip: Cyclic contrastive language-image pretraining.
\newblock {\em Advances in Neural Information Processing Systems}, 35:6704--6719, 2022.

\bibitem{dann}
Yaroslav Ganin and Victor Lempitsky.
\newblock Unsupervised domain adaptation by backpropagation.
\newblock In {\em International conference on machine learning}, pages 1180--1189. PMLR, 2015.

\bibitem{mmd}
Liwen Ouyang and Aaron Key.
\newblock Maximum mean discrepancy for generalization in the presence of distribution and missingness shift.
\newblock {\em arXiv preprint arXiv:2111.10344}, 2021.

\bibitem{meta3}
Qi~Dou, Daniel Coelho~de Castro, Konstantinos Kamnitsas, and Ben Glocker.
\newblock Domain generalization via model-agnostic learning of semantic features.
\newblock {\em Advances in neural information processing systems}, 32, 2019.

\bibitem{img-level1}
Kaiyang Zhou, Yongxin Yang, Timothy Hospedales, and Tao Xiang.
\newblock Deep domain-adversarial image generation for domain generalisation.
\newblock In {\em Proceedings of the AAAI conference on artificial intelligence}, volume~34, pages 13025--13032, 2020.

\bibitem{img-level2}
Ekin~D Cubuk, Barret Zoph, Jonathon Shlens, and Quoc~V Le.
\newblock Randaugment: Practical automated data augmentation with a reduced search space.
\newblock In {\em Proceedings of the IEEE/CVF conference on computer vision and pattern recognition workshops}, pages 702--703, 2020.

\bibitem{dsu}
Xiaotong Li, Yongxing Dai, Yixiao Ge, Jun Liu, Ying Shan, and Ling-Yu Duan.
\newblock Uncertainty modeling for out-of-distribution generalization.
\newblock {\em arXiv preprint arXiv:2202.03958}, 2022.

\bibitem{styleneophile}
Juwon Kang, Sohyun Lee, Namyup Kim, and Suha Kwak.
\newblock Style neophile: Constantly seeking novel styles for domain generalization.
\newblock In {\em Proceedings of the IEEE/CVF Conference on Computer Vision and Pattern Recognition}, pages 7130--7140, 2022.

\bibitem{padain}
Oren Nuriel, Sagie Benaim, and Lior Wolf.
\newblock Permuted adain: Reducing the bias towards global statistics in image classification.
\newblock In {\em Proceedings of the IEEE/CVF Conference on Computer Vision and Pattern Recognition}, pages 9482--9491, 2021.

\bibitem{ccfp}
Chenming Li, Daoan Zhang, Wenjian Huang, and Jianguo Zhang.
\newblock Cross contrasting feature perturbation for domain generalization.
\newblock In {\em Proceedings of the IEEE/CVF International Conference on Computer Vision}, pages 1327--1337, 2023.

\bibitem{aloft}
Jintao Guo, Na~Wang, Lei Qi, and Yinghuan Shi.
\newblock Aloft: A lightweight mlp-like architecture with dynamic low-frequency transform for domain generalization.
\newblock In {\em Proceedings of the IEEE/CVF Conference on Computer Vision and Pattern Recognition}, pages 24132--24141, 2023.

\bibitem{diagnosing}
Yuhui Zhang, Jeff~Z HaoChen, Shih-Cheng Huang, Kuan-Chieh Wang, James Zou, and Serena Yeung.
\newblock Diagnosing and rectifying vision models using language.
\newblock {\em arXiv preprint arXiv:2302.04269}, 2023.

\bibitem{clip-adapter}
Peng Gao, Shijie Geng, Renrui Zhang, Teli Ma, Rongyao Fang, Yongfeng Zhang, Hongsheng Li, and Yu~Qiao.
\newblock Clip-adapter: Better vision-language models with feature adapters.
\newblock {\em International Journal of Computer Vision}, 132(2):581--595, 2024.

\bibitem{tip}
Renrui Zhang, Rongyao Fang, Wei Zhang, Peng Gao, Kunchang Li, Jifeng Dai, Yu~Qiao, and Hongsheng Li.
\newblock Tip-adapter: Training-free clip-adapter for better vision-language modeling.
\newblock {\em arXiv preprint arXiv:2111.03930}, 2021.

\bibitem{gestur}
Byounggyu Lew, Donghyun Son, and Buru Chang.
\newblock Gradient estimation for unseen domain risk minimization with pre-trained models.
\newblock In {\em Proceedings of the IEEE/CVF International Conference on Computer Vision}, pages 4436--4446, 2023.

\bibitem{vl2v}
Louis H{\'e}madou, H{\'e}l{\'e}na Vorobieva, Ewa Kijak, and Frederic Jurie.
\newblock Beyond internet images: Evaluating vision-language models for domain generalization on synthetic-to-real industrial datasets.
\newblock In {\em Synthetic Data for Computer Vision Workshop@ CVPR 2024}.

\bibitem{rise}
Zeyi Huang, Andy Zhou, Zijian Ling, Mu~Cai, Haohan Wang, and Yong~Jae Lee.
\newblock A sentence speaks a thousand images: Domain generalization through distilling clip with language guidance.
\newblock In {\em Proceedings of the IEEE/CVF International Conference on Computer Vision}, pages 11685--11695, 2023.

\bibitem{promptstyler}
Junhyeong Cho, Gilhyun Nam, Sungyeon Kim, Hunmin Yang, and Suha Kwak.
\newblock Promptstyler: Prompt-driven style generation for source-free domain generalization.
\newblock In {\em Proceedings of the IEEE/CVF International Conference on Computer Vision}, pages 15702--15712, 2023.

\bibitem{coop}
Kaiyang Zhou, Jingkang Yang, Chen~Change Loy, and Ziwei Liu.
\newblock Learning to prompt for vision-language models.
\newblock {\em International Journal of Computer Vision}, 130(9):2337--2348, 2022.

\bibitem{dpl}
Xin Zhang, Shixiang~Shane Gu, Yutaka Matsuo, and Yusuke Iwasawa.
\newblock Domain prompt learning for efficiently adapting clip to unseen domains.
\newblock {\em Transactions of the Japanese Society for Artificial Intelligence}, 38(6):B--MC2\_1, 2023.

\bibitem{duprg}
Hongjing Niu, Hanting Li, Feng Zhao, and Bin Li.
\newblock Domain-unified prompt representations for source-free domain generalization.
\newblock {\em arXiv preprint arXiv:2209.14926}, 2022.

\bibitem{stylip}
Shirsha Bose, Ankit Jha, Enrico Fini, Mainak Singha, Elisa Ricci, and Biplab Banerjee.
\newblock Stylip: Multi-scale style-conditioned prompt learning for clip-based domain generalization.
\newblock In {\em Proceedings of the IEEE/CVF Winter Conference on Applications of Computer Vision}, pages 5542--5552, 2024.

\bibitem{dspl}
De~Cheng, Zhipeng Xu, Xinyang Jiang, Nannan Wang, Dongsheng Li, and Xinbo Gao.
\newblock Disentangled prompt representation for domain generalization.
\newblock In {\em Proceedings of the IEEE/CVF Conference on Computer Vision and Pattern Recognition}, pages 23595--23604, 2024.

\bibitem{spg}
Shuanghao Bai, Yuedi Zhang, Wanqi Zhou, Zhirong Luan, and Badong Chen.
\newblock Soft prompt generation for domain generalization.
\newblock {\em arXiv preprint arXiv:2404.19286}, 2024.

\bibitem{lora}
Edward~J Hu, Yelong Shen, Phillip Wallis, Zeyuan Allen-Zhu, Yuanzhi Li, Shean Wang, Lu~Wang, and Weizhu Chen.
\newblock Lora: Low-rank adaptation of large language models.
\newblock {\em arXiv preprint arXiv:2106.09685}, 2021.

\bibitem{hydra}
Chunlin Tian, Zhan Shi, Zhijiang Guo, Li~Li, and Chengzhong Xu.
\newblock Hydralora: An asymmetric lora architecture for efficient fine-tuning.
\newblock {\em arXiv preprint arXiv:2404.19245}, 2024.

\bibitem{dan}
Mingsheng Long, Yue Cao, Jianmin Wang, and Michael Jordan.
\newblock Learning transferable features with deep adaptation networks.
\newblock In {\em International conference on machine learning}, pages 97--105. PMLR, 2015.

\bibitem{jan}
Mingsheng Long, Han Zhu, Jianmin Wang, and Michael~I Jordan.
\newblock Deep transfer learning with joint adaptation networks.
\newblock In {\em International conference on machine learning}, pages 2208--2217. PMLR, 2017.

\bibitem{cocoop}
Kaiyang Zhou, Jingkang Yang, Chen~Change Loy, and Ziwei Liu.
\newblock Conditional prompt learning for vision-language models.
\newblock In {\em Proceedings of the IEEE/CVF conference on computer vision and pattern recognition}, pages 16816--16825, 2022.

\bibitem{fourier}
Qinwei Xu, Ruipeng Zhang, Ya~Zhang, Yanfeng Wang, and Qi~Tian.
\newblock A fourier-based framework for domain generalization.
\newblock In {\em Proceedings of the IEEE/CVF conference on computer vision and pattern recognition}, pages 14383--14392, 2021.

\bibitem{maple}
Muhammad~Uzair Khattak, Hanoona Rasheed, Muhammad Maaz, Salman Khan, and Fahad~Shahbaz Khan.
\newblock Maple: Multi-modal prompt learning.
\newblock In {\em Proceedings of the IEEE/CVF Conference on Computer Vision and Pattern Recognition}, pages 19113--19122, 2023.

\bibitem{vpt}
Menglin Jia, Luming Tang, Bor-Chun Chen, Claire Cardie, Serge Belongie, Bharath Hariharan, and Ser-Nam Lim.
\newblock Visual prompt tuning.
\newblock In {\em European Conference on Computer Vision}, pages 709--727. Springer, 2022.

\bibitem{vl2v-sd}
Sravanti Addepalli, Ashish~Ramayee Asokan, Lakshay Sharma, and R~Venkatesh Babu.
\newblock Leveraging vision-language models for improving domain generalization in image classification.
\newblock In {\em Proceedings of the IEEE/CVF Conference on Computer Vision and Pattern Recognition}, pages 23922--23932, 2024.

\bibitem{cliplora}
Maxime Zanella and Ismail Ben~Ayed.
\newblock Low-rank few-shot adaptation of vision-language models.
\newblock In {\em Proceedings of the IEEE/CVF Conference on Computer Vision and Pattern Recognition}, pages 1593--1603, 2024.

\bibitem{gumble}
Iris~AM Huijben, Wouter Kool, Max~B Paulus, and Ruud~JG Van~Sloun.
\newblock A review of the gumbel-max trick and its extensions for discrete stochasticity in machine learning.
\newblock {\em IEEE Transactions on Pattern Analysis and Machine Intelligence}, 45(2):1353--1371, 2022.

\end{thebibliography}

\end{document}